\documentclass{article}
\usepackage{multirow}

\usepackage{arxiv}
\usepackage[utf8]{inputenc} 
\usepackage[T1]{fontenc}    
\usepackage{hyperref}       
\usepackage{url}            
\usepackage{booktabs}       
\usepackage{amsfonts}       
\usepackage{nicefrac}       
\usepackage{microtype}      
\usepackage{lipsum}
\usepackage[bottom]{footmisc} 
\usepackage{graphicx}
\usepackage{amsmath}
\usepackage{array}
\usepackage{wasysym}
\usepackage{multirow}
\usepackage{adjustbox}
\usepackage{authblk}  
\title{Can Multi-modal (reasoning) LLMs work as deepfake detectors?}
\author[1]{Simiao Ren\textsuperscript{\dag}}
\author[3]{Yao Yao\textsuperscript{*}}
\author[4]{Kidus Zewde\textsuperscript{*}}
\author[5]{Zisheng Liang\textsuperscript{*}}
\author[4]{Tsang (Dennis) Ng\textsuperscript{*}}
\author[4]{Ning-Yau Cheng\textsuperscript{*}}
\author[3]{Xiaoou Zhan\textsuperscript{*}}
\author[6]{Qinzhe Liu\textsuperscript{*}}
\author[5]{Yifei Chen\textsuperscript{*}}
\author[2]{Hengwei Xu\textsuperscript{*}}

\affil[1]{Duke University, \texttt{simiao.ren/zisheng.liang@duke.edu}}
\affil[2]{Georgia Tech, \texttt{hxu457@gatech.edu}}
\affil[3]{University of Wisconsin Madison, \texttt{yyao39/xzhan23@wisc.edu}}
\affil[4]{Scam.ai, \texttt{\{dennis.ng/kidus.zewde/rachel.cheng\}@scam.ai}}
\affil[5]{Columnbia University, \texttt{yc3503@columbia.edu}}
\affil[6]{\texttt{qinzhe1126@gmail.com}}

\begin{document}
\maketitle
{\renewcommand{\thefootnote}{}\footnote{Authorship of this paper does not imply employment by or affiliation with Scam.AI (Reality Inc.); some authors are independent researchers who collaborated on this work.}\addtocounter{footnote}{-1}}

\begin{abstract}
Deepfake detection remains a critical challenge in the era of advanced generative models, particularly as synthetic media becomes more sophisticated. In this study, we explore the potential of state of the art multi-modal (reasoning) large language models (LLMs) for deepfake image detection such as (OpenAI O1/4o, Gemini thinking Flash 2, Deepseek Janus, Grok 3, llama 3.2, Qwen 2/2.5 VL, Mistral Pixtral, Claude 3.5/3.7 sonnet)  . We benchmark 12 latest multi-modal LLMs against traditional deepfake detection methods across multiple datasets, including recently published real-world deepfake imagery. To enhance performance, we employ prompt tuning and conduct an in-depth analysis of the models' reasoning pathways to identify key contributing factors in their decision-making process. Our findings indicate that best multi-modal LLMs achieve competitive performance with promising generalization ability with zero shot, even surpass traditional deepfake detection pipelines in out-of-distribution datasets while the rest of the LLM families performs extremely disappointing with some worse than random guess. Furthermore, we found newer model version and reasoning capabilities does not contribute to performance in such niche tasks of deepfake detection while model size do help in some cases. This study highlights the potential of integrating multi-modal reasoning in future deepfake detection frameworks and provides insights into model interpretability for robustness in real-world scenarios

\end{abstract}

\section{INTRODUCTION}

\renewcommand{\thefootnote}{\fnsymbol{footnote}}

\footnotetext[2]{\dag\ Correspondence author.}
\footnotetext[1]{* Equal contributions, order randomly generated.}

The emergence of artificial intelligence has yielded significant advantages across a diverse range of disciplines, encompassing material design and energy sectors\cite{ren2020benchmarking, ren2022automated, ren2024segment, mandal2020acoustic}. However, rapid advancement of generative models has led to the proliferation of highly realistic deepfake images, raising significant concerns about misinformation, identity fraud, and the erosion of digital trust. Traditional deepfake detection methods rely primarily on convolutional neural networks (CNNs) and vision-based deep learning models to identify artifacts and inconsistencies in synthetic images. However, as generative models improve, these methods struggle with generalization, particularly when faced with real-world deepfake manipulations that deviate from training data distributions \cite{ren2025deepfake}.

Recent progress in multi-modal large language models (LLMs) \cite{lu2024deepseek, jaech2024openai, agrawal2024pixtral, grattafiori2024llama, team2023gemini, yang2024qwen2, claude2025Anthropic} has demonstrated their ability to integrate and reason across multiple data modalities, including vision and text. These models, such as GPT-4V and Gemini, leverage vast pretraining \cite{team2023gemini} corpora and emergent reasoning capabilities, making them promising candidates for deepfake detection. Unlike conventional vision-based methods, multi-modal LLMs have the potential to incorporate contextual information, reason about image features, and adapt dynamically to unseen manipulations.

In this study, we systematically evaluate the effectiveness of multi-modal LLMs for deepfake detection. We benchmark several state-of-the-art (SOTA) models across multiple datasets, including real-world deepfake imagery, to assess their generalization capabilities. Furthermore, we conduct ablation studies and probe their reasoning processes to identify key contributing factors in their decision-making. Our analysis provides insights into how these models reason about deepfake images, shedding light on their interpretability and limitations.

Our contributions are threefold:

\begin{itemize}
    \item Benchmarking SOTA (up to Mar 2025) Multi-Modal LLMs on deepfake detection: We evaluate the performance of leading multi-modal LLMs against traditional deepfake detection methods on diverse datasets, including real-world deepfake samples, which were never fairly benchmarked on such LLMs and first paper featuring reasoning multi-modal LLMs on this task.
    \item Ablation studies into their reasoning capabilities \& Interpretability: We investigate the reasoning steps taken by multi-modal LLMs, identifying the key features and logical pathways that contribute to their decisions.
    \item  Our findings indicate that SOTA multi-modal LLMs exhibit promising generalization abilities and competitive performance compared to traditional detection approaches. This study provides valuable insights into the role of reasoning in deepfake detection and lays the groundwork for future research into robust, interpretable, and generalized detection systems.
\end{itemize}

\section{Related work}

\textbf{LLM on Deepfake Detection: Emerging Research and Findings}
Deepfake detection has traditionally relied on computer vision techniques \cite{yan2023deepfakebench}, such as convolutional neural networks \cite{shiohara2022detecting}, to identify anomalies in visual features like pixel-level inconsistencies \cite{DFDC2020}. However, with the rise of generative AI, particularly multi-modal LLMs, there is growing interest in leveraging these models for deepfake detection due to their semantic understanding and reasoning capabilities. These models can process both text and images, offering a novel approach to identifying AI-generated media, which is critical given the increasing sophistication of deepfakes and their potential for misuse in spreading disinformation.

A significant study in this area is Li et al. \cite{jia2024can} that investigates the capabilities of multimodal LLMs, specifically OpenAI’s GPT4V Vision model and Google Gemini 1.0 Pro from the FFHQ dataset and 2,000 AI-generated images from the DF dataset, generated by StyleGAN2 and Latent Diffusion models. The approach involved zero-shot prompting, where the LLMs were asked to determine if an image was real or AI-generated without additional training. Our work features signficantly more (12 compared to 2) and more advanced LLMs. VP et al. \cite{vp2024llm} combined multi-modal LLMs with the CNN features to predict deepfake and showed promising result with InstructBLIP model infusing with regular CNN structure, we are distinct with them as we only use the LLM with zero-shot in this work. Liu et al \cite{liu2024evolving} surveyed multiple papers that utilized multi-modal detection techniques that utilizes LLMs, however they focused on the paradigms and processes instead of the LLMs themselves, which is the focus of this work. 

\textbf{State of the Art Multi-Modal LLMs: Capabilities and Advances}
We list the LLMs that we benchmark 
\begin{enumerate}
    \item \textbf{OpenAI GPT-4o} \\
        \textit{Architecture and Features:} GPT-4o, released in May 2024, is a multimodal model capable of processing text, images, and audio, with a focus on natural human-computer interaction \cite{openai2024vision}. \\
        \textit{Performance Metrics:} Achieves 69.1\% on MMMU and 85.3\% on VQA v2, indicating strong performance in complex visual and reasoning tasks \cite{gpt4oexplained}. 
  \item \textbf{OpenAI GPT-o1} \\
        \textit{Architecture and Features:} GPT-O1, released in Sep 2024, is the first multimodal model that OpenAI published with reasoning capabilities \cite{openai2024vision}. \\
        \textit{Performance Metrics:} Achieves 78.2\% MMMU (the highest in our benchmark list) \cite{gpt4oexplained}. 
    \item \textbf{Google Gemini 2} \\
        \textit{Architecture and Features:} Part of Google's Gemini family released Dec 2024, it supports text, images, and video inputs, emphasizing real-time processing \cite{geminiapi}. It also has a reasoning version currently versioned as 'gemini-2.0-flash-thinking-exp-01-21' \\
        \textit{Performance Metrics:} Scores 68.5\% on MMMU and 84.5\% on VQA v2, closely competing with GPT-4o \cite{geminivision}. Its reasoning versioning scores a MMU of 75.4\% which is our 2nd performing one in this benchmarked list of LLMs. 
    \item \textbf{Deepseek Janus-Pro} \\
        \textit{Architecture and Features:} Released in Jan 2025. Features decoupled visual encoding for understanding and generation \cite{januspaper}. It has 7B and 1B two versions. We are benchmarking both of them in our model. \\
        \textit{Performance Metrics:} 1B: 36.3\% in MMMU while 7B has 41.0\% \cite{januspro}. 
    \item \textbf{Meta Llama 3.2 Vision} \\
        \textit{Architecture and Features:} Launched Sep 2024, llama 3.2 11B vision is one of the most powerful open-sourced vision foundational models. It builds on top of the llama 3.1. \cite{llama32vision}. \\
        \textit{Performance Metrics:} Scores 54.1\% on MMMU, excelling in AI2 Diagram (92.3) and DocVQA (90.1) \cite{llama32guide}. 
    \item \textbf{Alibaba Qwen 2/2.5 VL} \\
        \textit{Architecture and Features:} Released in Jan 2025, it has enhances vision-language integration with dynamic resolution ViT \cite{qwen25vlreport}. The 2.0 version was released in Aug 2024 and has 7B parameters. \\
        \textit{Performance Metrics:} Achieves 70.3\% on MMMU, strong in document understanding \cite{qwen25vl}. 
    \item \textbf{Mistral Pixtral} \\
        \textit{Architecture and Features:} Released in Nov 2024, it supports variable image sizes with a 12B model \cite{pixtral12b}. \\
        \textit{Performance Metrics:} Scores 52.5\% on MMMU, efficient for its size \cite{pixtral12bmedium}. 
    \item \textbf{Anthropic Claude 3.5 Haiku / 3.7 Sonnet} \\
        \textit{Architecture and Features:} 3.5 Haiku was released on Nov 2024 while 3.7 was released on Feb 2025. They offers vision capabilities across the Claude 3 family \cite{claudevision}. \\
        \textit{Performance Metrics:} Scores 50.2\% on MMMU aand 3.7 has 71.8\% \cite{claudegoogle}.
\end{enumerate}

\begin{table}[h]
    \centering
    \caption{Comparative Vision Performance of Multi-Modal LLMs}
    \begin{tabular}{llcc}
        \toprule
        \textbf{Publisher} & \textbf{Model Version} & \textbf{MMMU (\%)} & \textbf{VQA v2 Acc (\%)} \\
        \midrule
        \multirow{2}{*}{OpenAI} & GPT-4o & 69.1 & 85.3 \\
                                & GPT-o1 & 78.2 & N/A \\
        Meta                    & Llama 3.2 & 71.3 & 78.1 \\
        \multirow{2}{*}{Google} & Gemini 2 & 68.5 & 84.5 \\
                                & Gemini 2 Thinking & 75.4 & N/A \\
        \multirow{2}{*}{Alibaba}& Qwen 2 VL & 45.2 & N/A \\
                                & Qwen 2.5 VL & 70.3 & N/A \\
        \multirow{2}{*}{Anthropic} & Claude 3.5 Haiku & 50.2 & N/A \\
                                   & Claude 3.7 Sonnet & 71.8 & N/A \\
        \multirow{2}{*}{Deepseek}  & Janus-Pro 1B & 36.3 &  N/A \\
                                    & Janus-Pro 7B & 41.0 &  N/A \\
        Mistral                 & Pixtral & 52.5 & N/A \\
        \bottomrule
    \end{tabular}
\end{table}

\addtolength{\textheight}{-3cm}   

\section{Methodology}

\subsection{Overall Experiment design}

\begin{figure}[h]
    \centering
    \includegraphics[width=0.4\textwidth]{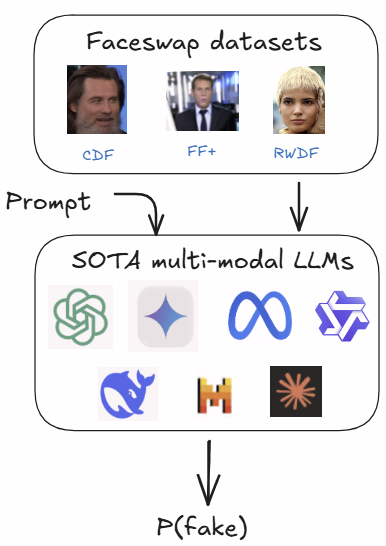}
    \caption{Overall experiment design}
    \label{fig:design}
\end{figure}

This section outlines the methodology employed to investigate the efficacy of state-of-the-art (SOTA) LLMs in detecting deepfake images using a structured pipeline, as depicted in the provided diagram \ref{fig:design}. The pipeline leverages faceswap datasets, processes them through SOTA multi-modal LLMs with tailored prompts, and generates a probability of an image being fake, denoted as P(Fake) in a zero-shot manner. The following subsections detail each step of the process. 

\subsection{ Faceswap Datasets}
The methodology begins with the selection and preparation of faceswap datasets, which serve as the foundation for evaluating the multi-modal LLMs. Three specific datasets are utilized:

CDF (Celeb-DeepFake Dataset) \cite{li2020celeb}: This dataset contains a collection of real and synthetically manipulated face images, featuring individuals such as the example shown in the diagram, providing a baseline for deepfake detection.
FF+ (FaceForensics++ Dataset) \cite{rossler2019faceforensics++}: An extended version of the FaceForensics dataset, FF+ includes a diverse set of manipulated videos and extracted frames, exemplified by the suited individual.
RWDF (Real-World DeepFake Dataset) \cite{ren2025deepfake}: This dataset comprises deepfake images from real-world sources, such as the blonde-haired individual depicted, capturing the nuances of naturally occurring deepfakes encountered in practical settings.
These datasets are preprocessed to ensure consistency in image resolution, format, and labeling, enabling a standardized input for the subsequent stages of the pipeline. We randomly sample 500 real and 500 fake images equally from each of the three dataset for form our evaluation set, this is due to the high cost of API calls to the most advanced LLMs.

\subsection{ Prompt Engineering and Input to SOTA Multi-Modal LLMs}
The preprocessed images from the faceswap datasets are fed into a suite of SOTA multi-modal LLMs, which are capable of processing both visual and textual data. The input process is guided by carefully crafted prompts designed to elicit reasoning and classification capabilities from the models. The prompt, as indicated in the diagram, directs the LLMs to analyze the input images and determine their authenticity. After some iterations we landed on the below prompt: "Is this image a deepfake or faceswap? Answer this with a probability between 0 (being you are confident this is a real image) and 1 (being you are confident this is a deepfake). Only answer one number with no other output". Each model receives the same prompt, ensuring consistency across evaluations. 

\subsection{ Evaluation and Analysis}
To assess the performance of the pipeline, the P(Fake) scores generated by each SOTA multi-modal LLM are benchmarked against ground truth labels from the faceswap datasets. Since language models does not necessarily returns structured outputs, we process the output with string processors to extract the numeric answer of the model. Note that this process does not guarantee full recovery of the prediction score, combined with a small probability that LLMs refuse to give a numeric estimation or just fails to reply. We omit those examples in our result calculation. 

For the evaluation metrics, we use the ROC curves with area under the receiver operating characteristic curve (AUC), as per countless previous literature \cite{ren2025deepfake, yan2023deepfakebench, deepfake-detection-challenge}. Additionally, an in-depth analysis of the reasoning pathways is conducted to identify the key contributing factors. 

\subsection{Ablation studies on what contributes to model performance and do they hallucinate}
We conduct multiple ablation studies or targeted comparisons to answer specific research questions. With the amount of SOTA LLMs we benchmark, we would like to investigate whether model version \/ model size \/ reasoning \/ would affect the performance of LLMs on this niche usecase of deepfake detection. 

To detect whether the model hallucinate, we higher the temperature and try the model with the exact same output multiple times to measure whether model is consistently outputing close results or its just randomly guessing with large variance.

\subsection{Benchmark traditional computer vision models}
To further understand and benchmark whether the LLMs can perform, we take two well-recognized and highly-performing benchmarked traditional Computer Vision models that were in multiple prior publications. The first model we chose self-blended network \cite{chen2022self} that was trained on FF+ dataset with an crafted training technique that does not rely on the fake images. The second model we chose is a naive efficient-net b4 model that was trained in a traditional supervised manner on FF+ dataset and performed extremely well in \cite{yan2023deepfakebench}. Both models were also further benchmarked in subsequent publications \cite{ren2025deepfake, yan2023deepfakebench} and are perfectly reimplemented (as we completely reused the code) in this study.

\begin{figure*}[h!]
    \centering
    \includegraphics[width=0.8\textwidth]{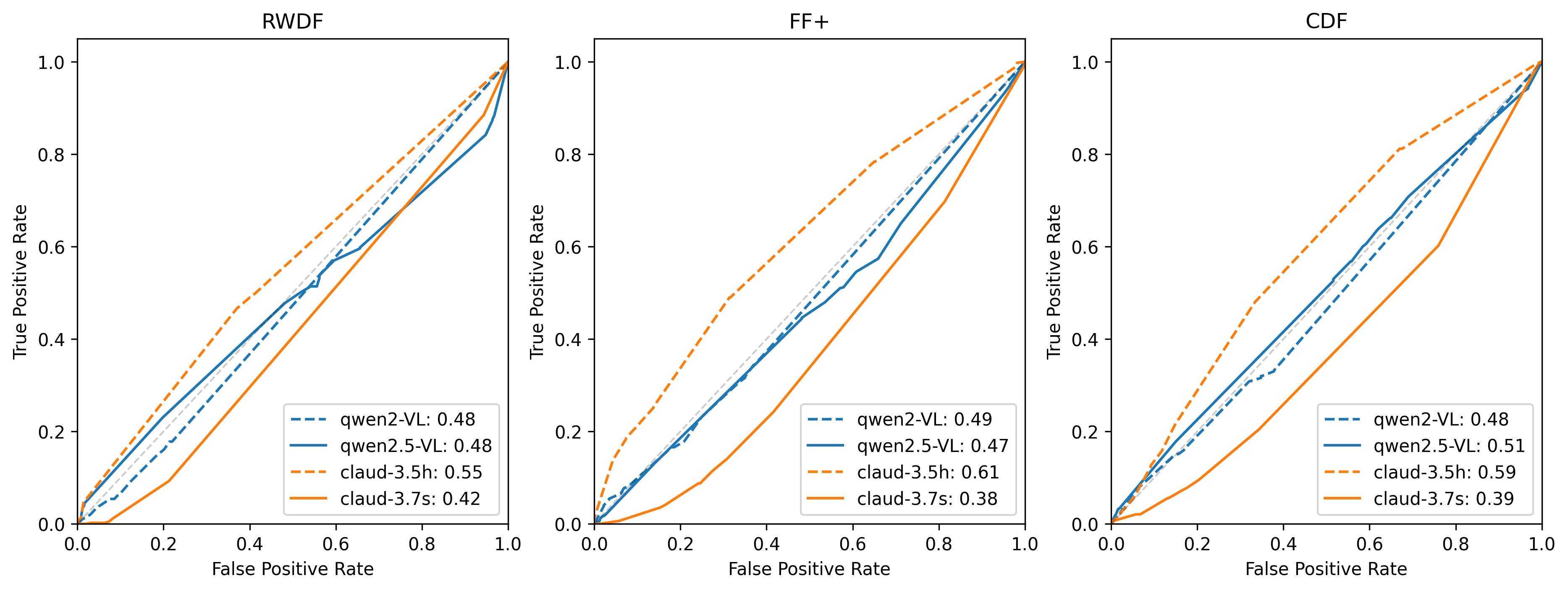}
    \caption{Version ROC Curve}
    \label{fig:version_roc}
\end{figure*}

\section{Results}
\subsection{Comparison between different LLMs}

\subsection{Does general model improvement (newer versions) generalizes to niche applications like deepfake>}

We are comparing newer versions (in time) performances. To control for the model architecture, data used and training hacks, we control the model publisher. In our benchmarked models, we have below pairs that we can gauge whether newer models has improvements on our niche task of deepfake.

\begin{itemize}
    \item Claud 3.5 Haiku VS Claude 3.7 Sonnet
    \item Qwen-2-VL-7B-Instruct VS Qwen-2.5-VL-7B-Instruct
\end{itemize}

Results comparison is shown in the figure \ref{fig:version_roc} shows that the AUROC measure did not increase with the model update from Qwen and Claud, with both of them performing as or even lower than the random guess. This shows that while newer model refresh performs significantly better on other benchmarks, their generalization ability to other niche tasks like deepfake detection is not improved, at least in these two models scenario.

\begin{figure*}[h!]
    \centering
    \includegraphics[width=0.8\textwidth]{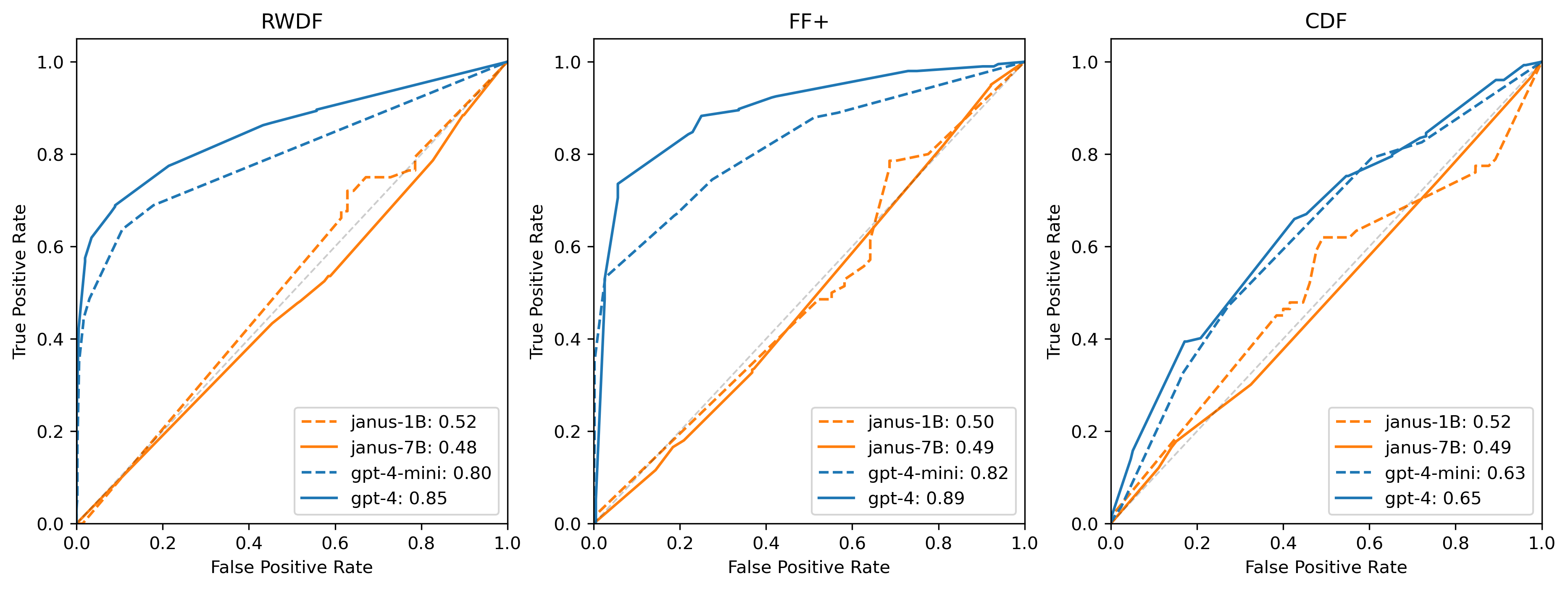}
    \caption{Size difference ROC Curve}
    \label{fig:size_roc}
\end{figure*}

\subsection{Does model size (parameter size) correlate with niche application like deepfake?}

We are comparing larger models in terms of number of trainable parameters. To control for the model architecture, data used and training hacks, we control the model publisher. In our benchmarked models, we have below pairs that we can gauge whether newer models has improvements on our niche task of deepfake.

\begin{itemize}
    \item DeepSeek-AI-Janus-Pro-1B VS DeepSeek-AI-Janus-Pro-7B 
    \item GPT-4o-Mini VS GPT-4o
\end{itemize}

Results comparison is shown in the figure \ref{fig:size_roc} shows that the AUROC measure do improve for GPT-4o model. Its AUROC improved 2-7\% in all three benchmarked tasks consistently, showing the number of parameters indeed help the model make more generalizable to decision on niche tasks. However, we also see that deepseek Janus model performed close to the random guess curve, despite the 7 times more parameter with its larger model. We will explore its failure modes in the below section.

\subsubsection{Do reasoning models perform better?}

We are comparing whether reasoning models that are trained on RL techniques help models generalize to better performance even for completely unseen niche task like deepfaked detection. To control for the model architecture, data used and training hacks, we control the model publisher. In our benchmarked models, we have below pairs that we can gauge whether newer models has improvements on our niche task of deepfake.

\begin{itemize}
    \item Gemini-2.0-Flash  VS Gemini-2.0-Flash-Thinking
    \item OAI-o1 VS GPT4o 
\end{itemize}

\begin{figure*}[h!]
    \centering
    \includegraphics[width=0.8\textwidth]{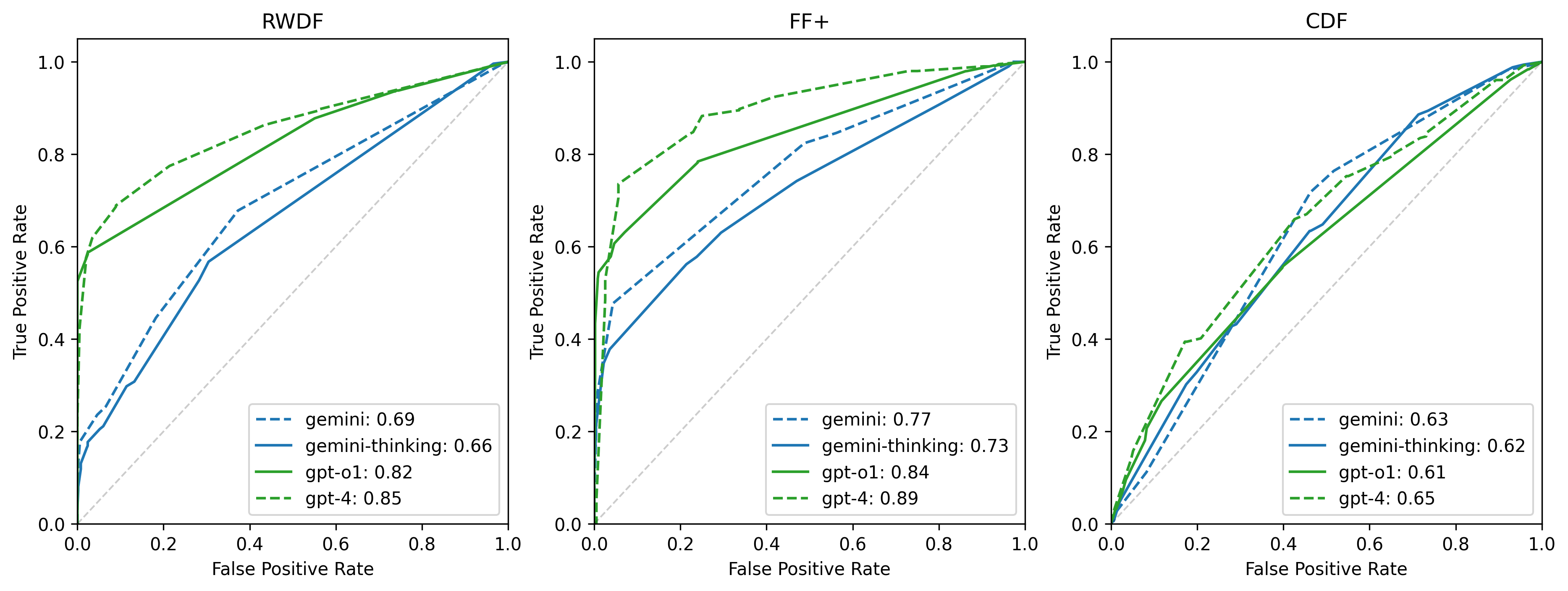}
    \caption{Reasoning ROC Curve}
    \label{fig:reasoning_roc}
\end{figure*}

Results comparison is shown in the figure \ref{fig:reasoning_roc} shows the comparison between reasoning models (solid lines) and their counter parts (dashed lines). We can see that reasoning not only did not bring any performance improvement in our task to detect deepfakes, but also introduced performance degradation. This is extremely interesting given the numerous reports \cite{guo2025deepseek, jaech2024openai} for reasoning's ability to generalize to more unseen tasks. We explore the potential reason why reasoning is performing much worse than even random chance in below section.

\subsection{Overll performance}
\begin{figure*}[h!]
    \centering
    \includegraphics[width=0.8\textwidth]{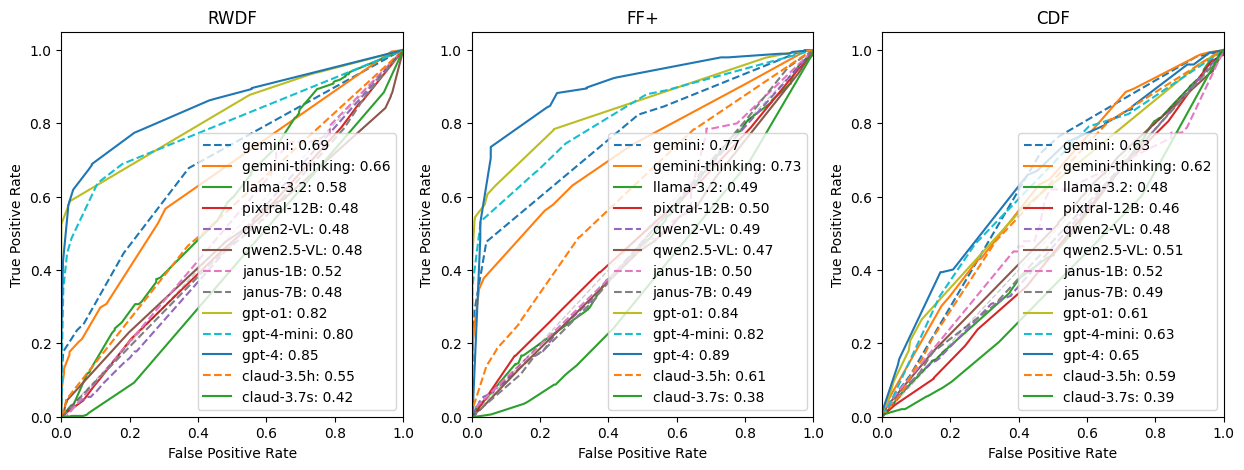}
    \caption{ROC Curve for all models}
    \label{fig:all_roc}
\end{figure*}

The overalll performance can be seen in the Figure \ref{fig:all_roc}. Frankly only openai models have reasonably well performance on our niche deepfake detection task while all other models struggles to have any real distinguishing power. 

\subsection{Do they hallucinate?}
\begin{figure}[h!]
    \centering
    \includegraphics[width=0.4\textwidth]{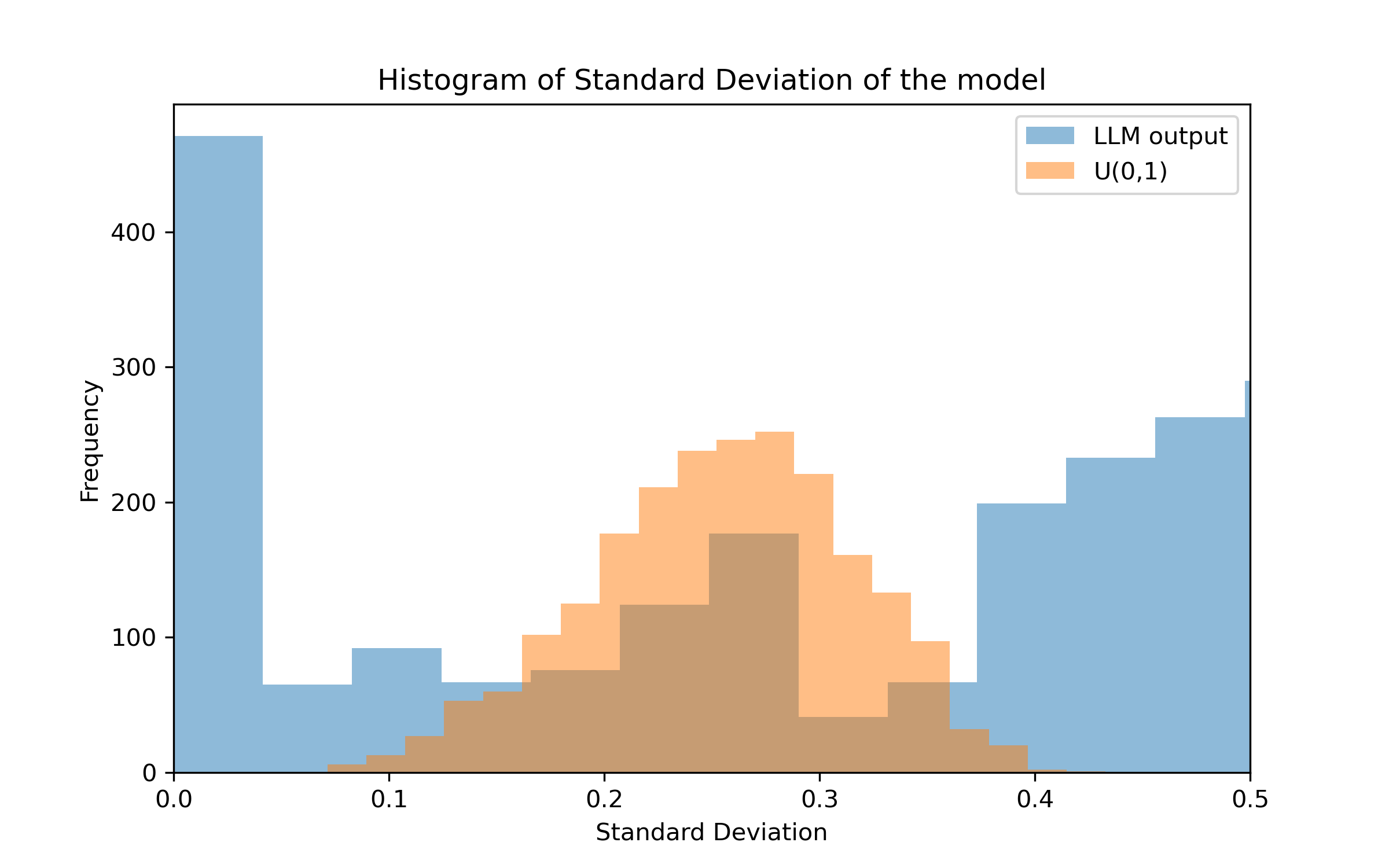}
    \caption{Standard deviation histogram of model output}
    \label{fig:std_hist}
\end{figure}

One core question we would like to answer is whether they hallucinate. Although the precise definition of "model hallucination" in non-factual tasks are not rigourously defined, seeing multiple of the multi-modal LLms performing close to the random guess line, we would like to explore the definition of hallucination of whether the model is confidently wrong or just randomly guessing. To achieve this, we asked the model the same question with a high temperature setting for multiple times and collected their standard deviation on the prediction P(fake). Due to budget limit we only tested this on Deepseek model. The result of this is shown in Figure \ref{fig:std_hist} where the model's output probability standard deviation is plot together with a uniform random distribution between 0-1 as a reference. From the plot we can see that the model is not randomly guessing but have 2 modes of images. A good portion of images have very small standard deviation, meaning the model is fairly consistent on its belief of whether this is deepfake or not while another big chunk of images shows the opposite behavior, outputing drastically different scores in the prediction. This shows that LLMs are "half randomly guessing", with half of its response highly varied while sticking to its answer in the other half of the cases.

\subsection{How did Open AI model performed so well? }

\begin{figure}[h!]
    \centering
    \includegraphics[width=0.4\textwidth]{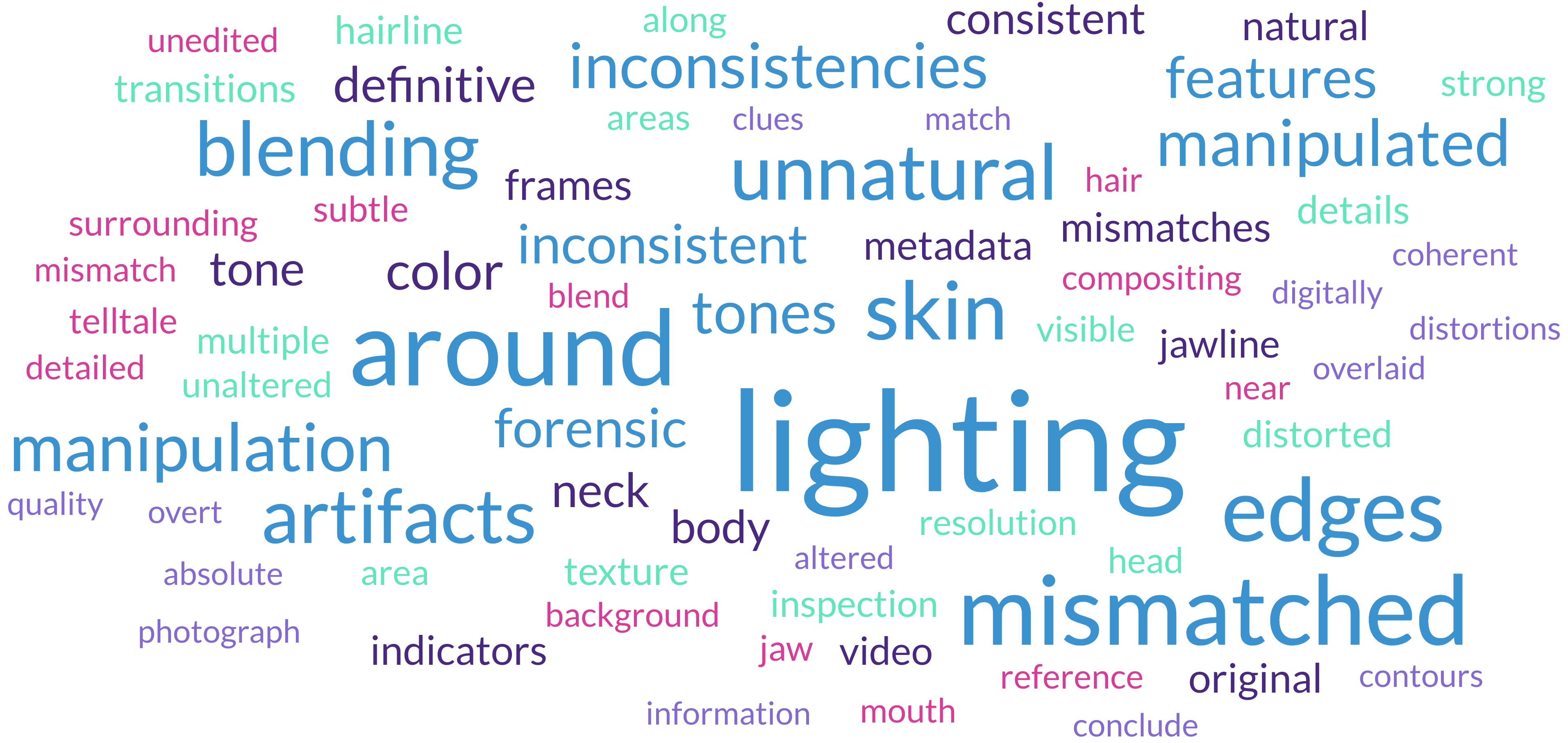}
    \caption{Word cloud from Open AI predictions}
    \label{fig:word_cloud}
\end{figure}

While explanability of LLMs is still an ongoing research, we did notice OpenAI models focus on a few themes in their analysis. Figure \ref{fig:word_cloud} shows a word cloud generated from the text output.

\begin{itemize}
    \item Discrepancies in lighting conditions between facial regions and the surroundings. A mismatch in lighting intensity, direction, or shadows across the face, neck, and body strongly influences the model’s judgment of manipulation.
    \item Differences in skin color between facial regions and adjacent body parts trigger suspicion. The model routinely highlights unnatural transitions, color disparities, or tonal inconsistencies around facial boundaries, particularly jawlines, hairlines, and neck areas. This is also why the word "around" appears in the word cloud.
    \item Visible artifacts such as abrupt or unnatural blending, slight blurring, misalignment at boundaries of facial features, or anomalous sharpness.
    \item Variations in texture clarity or resolution between the face and body regions.
\end{itemize}

In addition, Open AI model acknowledges limitations inherent in single-image evaluations, emphasizing that definitive forensic judgments typically require multiple reference images, video frames, higher-resolution imagery, or metadata analysis.

Quoted output for positive prediction:

\textit{"I see signs of possible blending and mismatches in skin tone and lighting around the jawline and hairline area, which suggest that the face region may have been composited from different sources. These inconsistencies commonly occur in images that have undergone significant facial manipulation."}

Quoted output for negative prediction:

\textit{"I see consistent lighting, focus, and color transitions around the face and neck with no obvious blending edges or distortions that often appear in deepfakes or faceswaps, so based on a visual check alone, it seems much more likely to be a real photo rather than a manipulated image."}

\subsection{What are they thinking: Failure mode analysis}

Through our investigate of the thinking pattern of LLMs, We found that the gpt model produced higher scores (above 0.5) for both categories in the CDF and FF++ datasets (score histogram can be found in appendix), while it produced lower scores (below 0.5) for both categories in the RWDF dataset. Next, we obtained the key characteristics that influenced model predictions and discovered that the model focused on aspects of image quality such as texture, blur, symmetry, background distortions, and the edges of facial features (for example, eyes and mouth). With this knowledge in mind, we further reviewed the three datasets and identified clear quality differences: images from CDF and FF++ are generally low quality and blurred, regardless of category, and the fake category in FF++ includes many highly distorted examples. In contrast, the images from RWDF are of higher quality.

These quality differences led us to infer that the varying performance of 4o across the three datasets may be attributable to differences in image quality. We hypothesize that although 4o is a reasonably effective image classifier to assess image quality, it has not been fine-tuned for the deep face detection task. With this hypothesis in mind, we examined the score distributions for the three datasets, See Figure \ref{fig:gpt_40}, and here are our findings. RWDF: The scores for real images (labeled as 0) are concentrated toward 0, while the scores for fake images (labeled as 1) are relatively evenly distributed between 0 and 1. This distribution suggests that real images in RWDF are reliably identified as real, whereas fake images span a range from clearly fake to nearly real. FF++: The score distribution for real images indicates that many of them are low quality and appear somewhat fake, while fake images are clearly identified as fake. This makes FF++ a relatively easy dataset to classify. CDF: The score distributions for real and fake images overlap considerably, with no distinct peak near 0 for real images. This suggests that real images in CDF are of relatively low quality and include many examples that appear fake, while fake images tend to be of higher quality and resemble real images—making CDF a comparatively challenging dataset.

\begin{figure}[h!]
\centering
\begin{tabular}{llll}
real & \includegraphics[width=.2\linewidth]{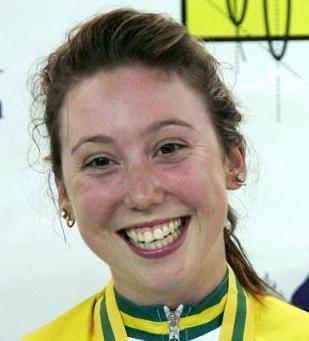} & \includegraphics[width=.2\linewidth]{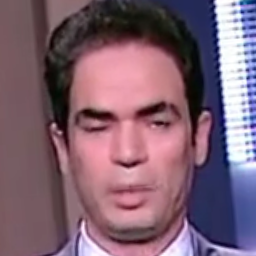} & \includegraphics[width=.2\linewidth]{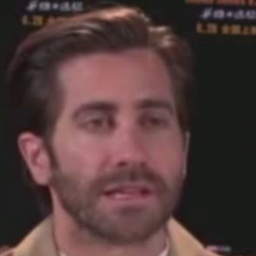}\\
fake & \includegraphics[width=.2\linewidth]{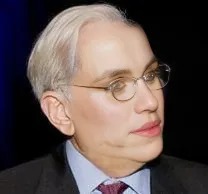} & \includegraphics[width=.2\linewidth]{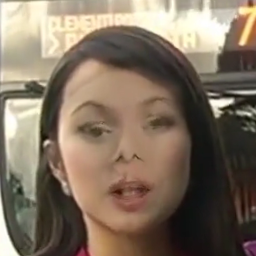} & \includegraphics[width=.2\linewidth]{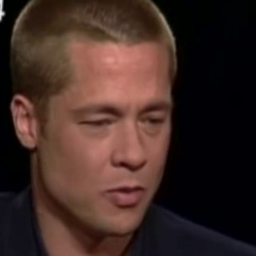}\\
\end{tabular}
\caption{Examples from the three datasets: RWDF (left), FF+ (middle), CDF (right). Top: real samples; bottom: deepfake counterparts.}
\label{fig:examples}
\end{figure}

These findings from the score distributions are consistent with our qualitative assessment of the three datasets, as illustrated in the examples below \ref{fig:examples}.

\subsubsection{Gemini thinking deepdive}

Research examining the Gemini 2.0 Flash model's performance in distinguishing real images from deepfakes highlighted both strengths and persistent weaknesses. Initially, using a simple prompt, the model displayed inconsistent confidence levels (0.25-0.99) and a heavy reliance on subjective factors, such as celebrity recognition (e.g., Brad Pitt influencing scores bidirectionally: 0.95-0.99 for deepfakes, 0.03-0.05 for real images), alongside image quality biases associating lower quality with authenticity. It detected features like oversmoothed skin (0.62-0.8) and lighting inconsistencies (0.75), but its methodology varied, leading to false positives for real images (e.g., 0.6-0.8) and an overreliance on contextual cues over technical analysis. A tuned prompt introduced a structured, technical framework—focusing on edge integrity, texture consistency, lighting, and resolution—eliminating identity bias and improving methodological consistency. This resulted in more balanced probability scores (0.2-0.75 for fakes, 0.1-0.7 for real) and detailed analyses, yet detection accuracy remained limited. While the tuned prompt mitigated biases and enhanced analytical rigor, the model’s fundamental discriminative ability showed little improvement, suggesting that its visual analysis limitations (like alignment between image and text embedding space), rather than prompt design alone, constrain effective deepfake detection, necessitating further model refinement.

For Gemini thinking (reasoning model), with the original prompt, the model exhibited a scoring inversion issue, assigning lower probabilities (0.1-0.3) to images it identified as deepfakes and higher scores (0.7-0.8) to those deemed real, contrary to the prompt's instructions where 0 indicated authenticity and 1 indicated manipulation. This inconsistency persisted despite detailed reasoning focusing on visual cues like skin texture, lighting, and edge blending. This is hypothesized to be the huge pre-training bias from the model overrides the prompt that explicitly expressed the score representation. A tuned prompt was then implemented, enforcing a structured, technical analysis of features such as edge integrity, texture consistency, and lighting, explicitly directing the model to ignore subject identity. This adjustment improved scoring consistency, particularly for real images (typically scored 0.1), and deepened the technical analysis, with notable sensitivity to beard texture and edge definition. However, detection of known deepfakes remained limited, with most assigned low probabilities (0.1-0.2), suggesting that while structured prompting enhanced analytical rigor, the model's underlying visual processing capabilities struggled to accurately identify manipulations. Possible reasons for the scoring inversion include prompt misinterpretation, training biases, or working memory limitations, though the persistence across both prompts indicates a deeper issue in the model’s detection framework rather than solely prompt design.

\begin{figure*}[h!]
    \centering
    \includegraphics[width=0.8\textwidth]{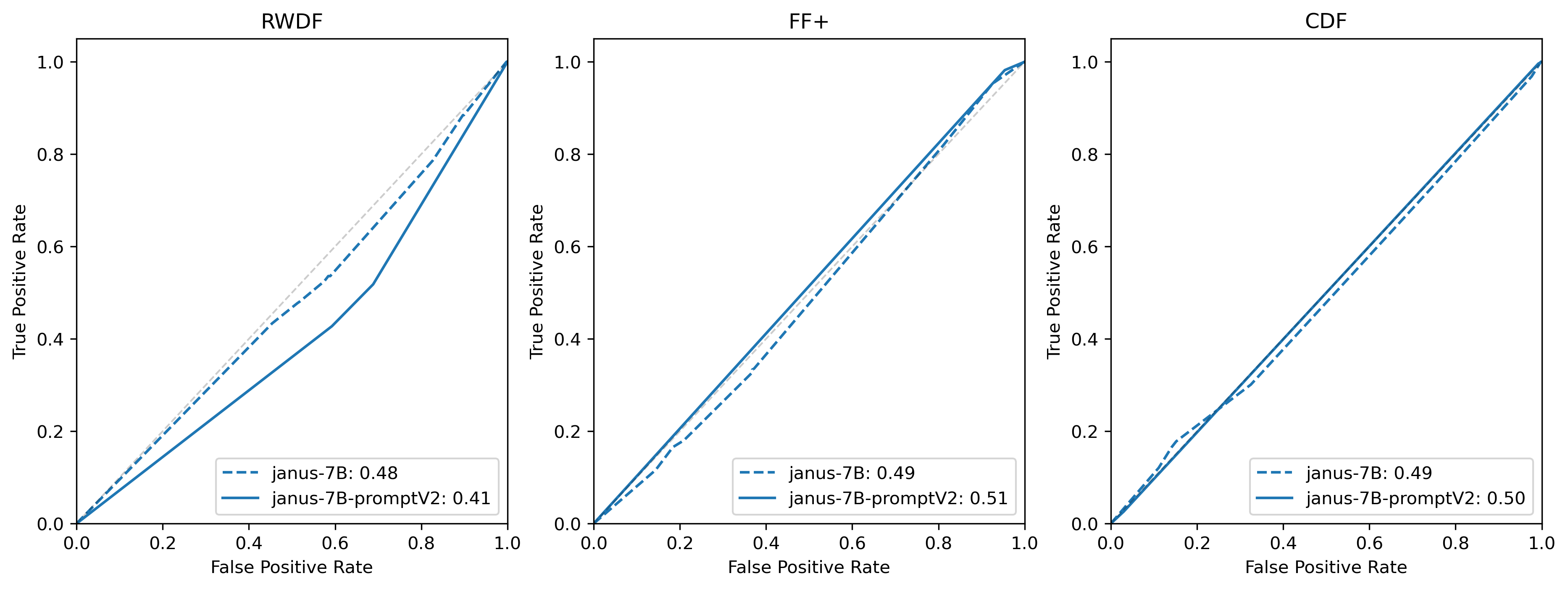}
    \caption{Float output vs categorical output}
    \label{fig:v2_prompt}
\end{figure*}

\begin{figure*}[h!]
    \centering
    \includegraphics[width=0.8\textwidth]{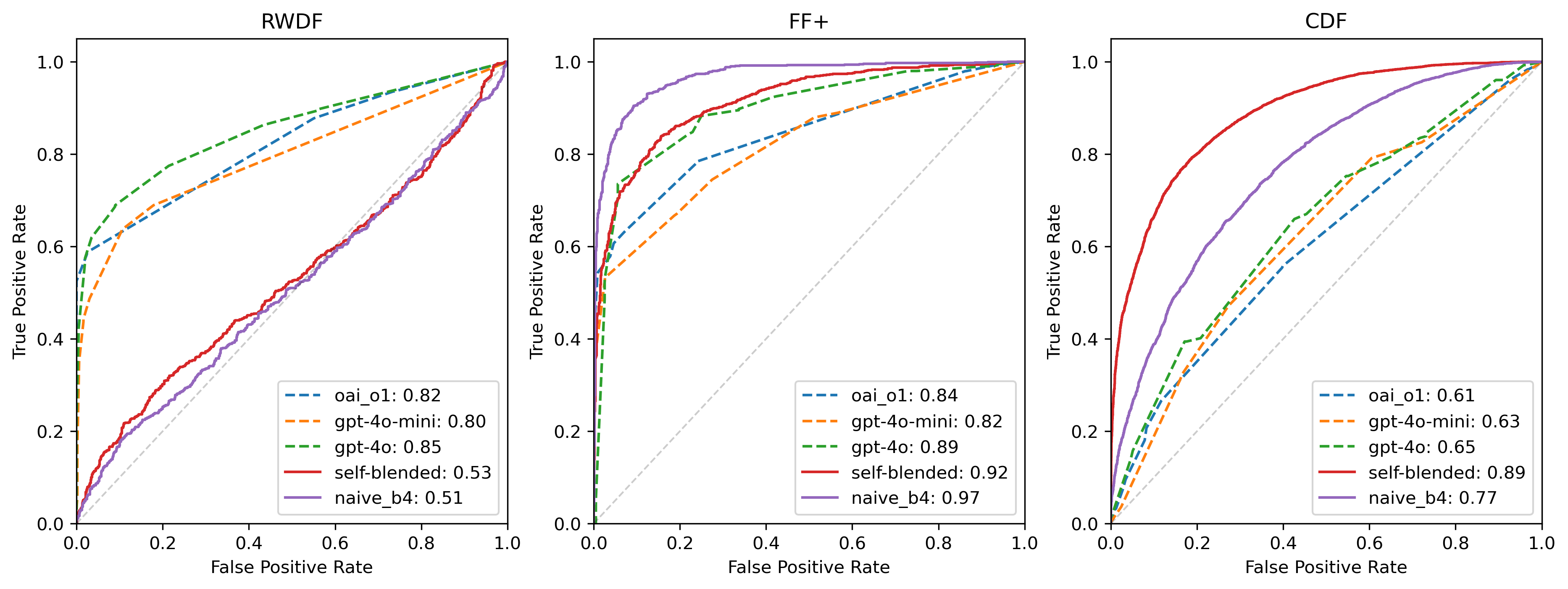}
    \caption{Comparing with traditional computer vision models that was trained on deepfake detection task}
    \label{fig:best_LLM_compare_traditional}
\end{figure*}

\subsection{How about asking binary classification instead of precise score prediction?}
One potential issue with LLMs is their weak perception of numeric relationship (like in the famous example of it struggling with 9.11 < 9.2). This is due to the tokenization nature of the LLMs and the distribution of the training data (reasoning models are usually much better on this). To ensure the model is not bottlenecked by this ability (i.e. model understand and can distinguish deepfakes, but they are so bad at outputing the probability because they are just bad at numbers), we also tested the prompt that eliminates this characteristics so that model only need to reply "real / fake / unsure". We scored them using "0/1/0.5" to generate the ROC curve. The result comparison is shown in \ref{fig:v2_prompt} where we see that the issue with deepseek unable to classify the deepfake not resulting from LLM's character of inability to make sense of the floating point number.

\subsection{LLMs comapred with traditional computer vision solutions}
We further compare them with traditional computer vision models that were trained on the deepfake detection tasks with tons of examples of real/fake and plot the top 3 performing LLMs (which all came from OpenAI) and visualize the results in figure \ref{fig:best_LLM_compare_traditional}. The best LLMs are dashed and the traditional methods are solid lines. We can see that the best LLMs from OpenAI are actually very generalized in terms of their ability to classify deepfake, under zero-shot settings (we hypothesize given deepfake detection is too niche, LLMs are very unlikely to have been trained specifically on this task). Our best mluti-modal LLMs consistently performs above random guessing while our traditional model overfit to their training set and fails to generalize into real-world settings \cite{ren2025deepfake}. That being said, on datasets that traditional networks are trained, they are still significantly better than even our best multi-modal LLMs.

\section{Conclusions}
This study demonstrates that state-of-the-art multi-modal large language models (LLMs) with reasoning capabilities offer promising potential for deepfake image detection, achieving competitive performance and notable generalization compared to traditional methods. Our benchmarking reveals that while model size can enhance performance, as seen with GPT-4o’s consistent improvements, newer model versions and reasoning-augmented approaches do not universally translate to better outcomes in this niche task, with some models (e.g., Claude, Qwen, and reasoning variants) performing near or below random guessing. OpenAI’s LLMs stood out, surpassing traditional computer vision models in zero-shot generalization, though they remain outperformed by specialized networks on datasets these were trained on. Additionally, our analysis suggests that poor performance is not solely attributable to numerical output limitations, as categorical prompting yielded similar results. These findings underscore the value of integrating multi-modal reasoning into deepfake detection frameworks while highlighting the need for further exploration into failure modes and model hallucination to enhance robustness and interpretability in real-world applications.

\section{Ethical impact statement}

The authors have read through the "Ethical Impact Statement Guidelines document" and agree to provide the below ethical impact statements:

It is clear that all studies and procedures described in the paper were not required to be approved by an ethic review board. In terms of potential hard to human subjects, as our paper does not involve any of the human nor animal subjects, there is no such potential harm foreseeable by authors. For the potential negative social impacts, the authors agree that the subject of deepfake and its detection or prevention is an active debating area in AI ethics and shall be dealt with great cautious. This work stands along side the harm prevention team where we try to help detect the real-world deepfakes, where those deepfakes can have significant negative potential impacts like unwillingly pornography or even jurisdiction falsehood. Due to the potential mitigation impact towards these adverse impacts, the authors believe that this work has a significant positive impact. 

\section{Appendix}

In this section we talk about things that are too detailed to be put in the main paper section to provide more context on our research journey.

\subsection{Histogram of each model output}
\begin{figure*}[h]
    \centering
    \includegraphics[width=\textwidth]{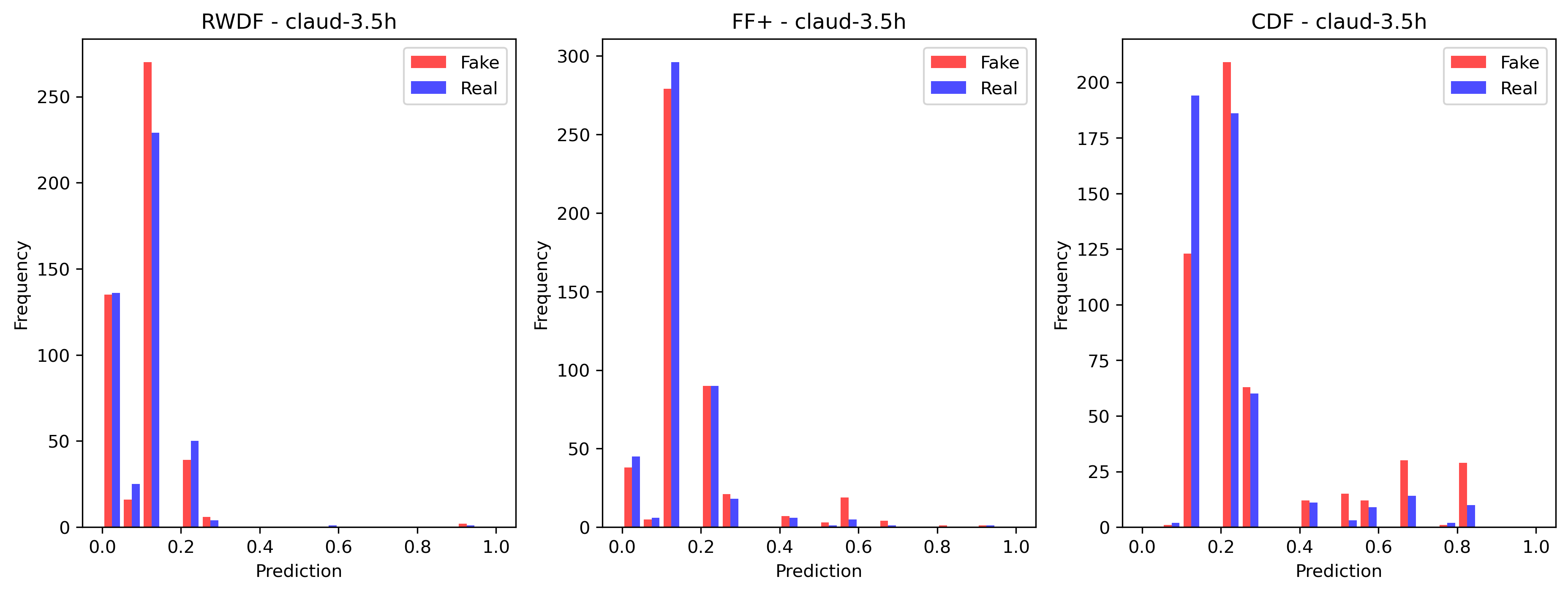}
    \caption{Claud 35 Haiku Histogram}
    \label{fig:claud_35_haiku}
\end{figure*}

\begin{figure*}[h]
    \centering
    \includegraphics[width=\textwidth]{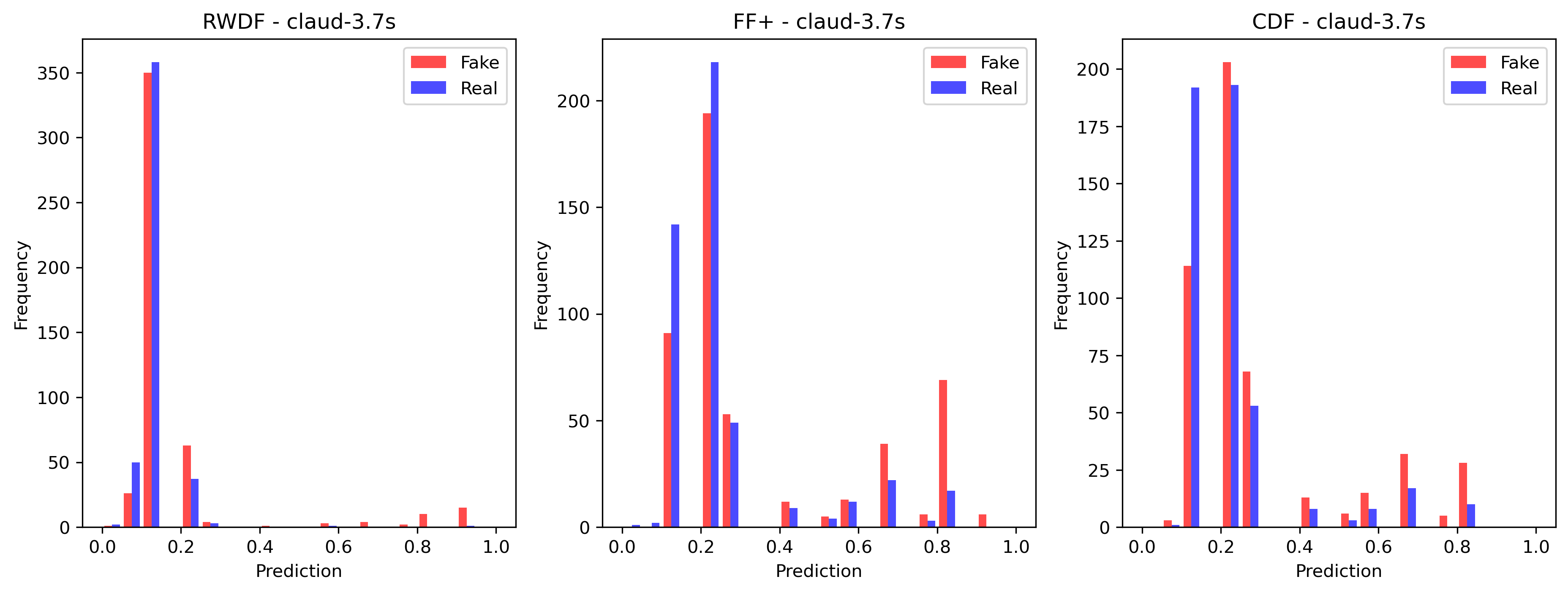}
    \caption{Claud 37 Sonnet Histogram}
    \label{fig:claud_37_sonnet}
\end{figure*}

\begin{figure*}[h]
    \centering
    \includegraphics[width=\textwidth]{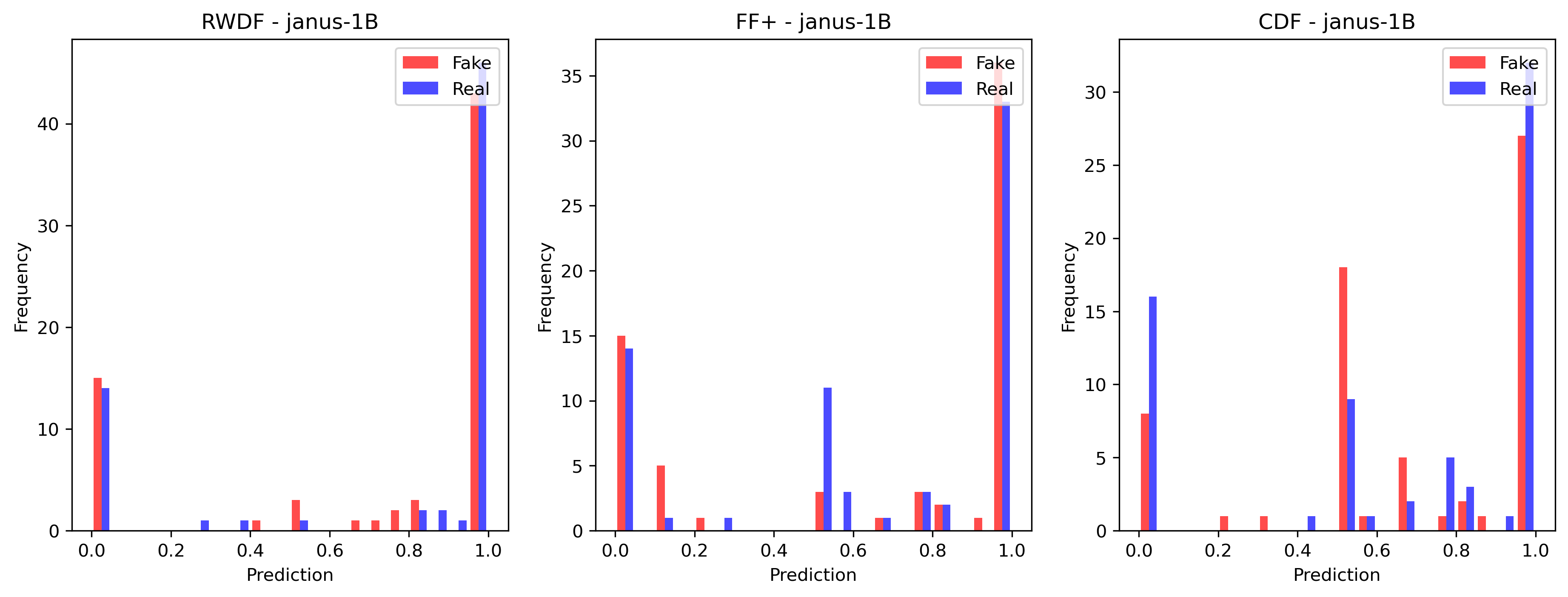}
    \caption{Deepseek AI Janus Pro 1B Histogram}
    \label{fig:deepseek_1b}
\end{figure*}

\begin{figure*}[h]
    \centering
    \includegraphics[width=\textwidth]{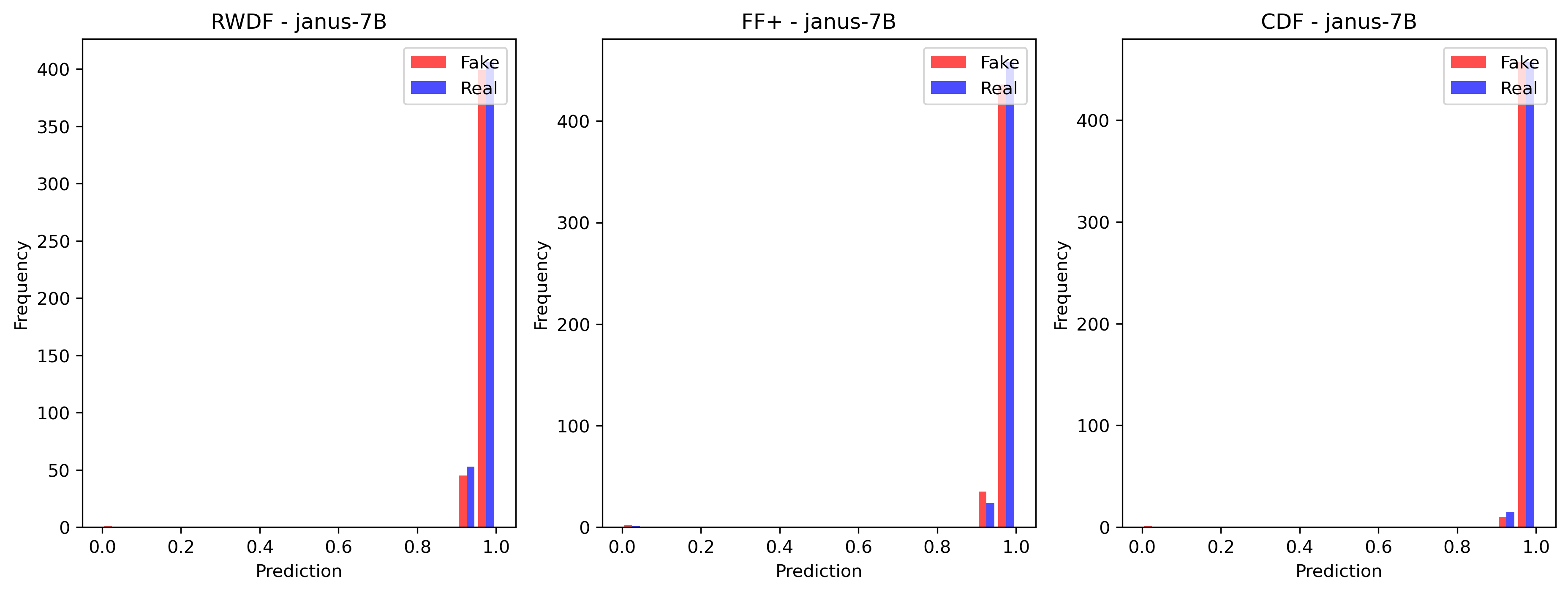}
    \caption{Deepseek AI Janus Pro 7B Histogram}
    \label{fig:deepseek_7b}
\end{figure*}

\begin{figure*}[h]
    \centering
    \includegraphics[width=\textwidth]{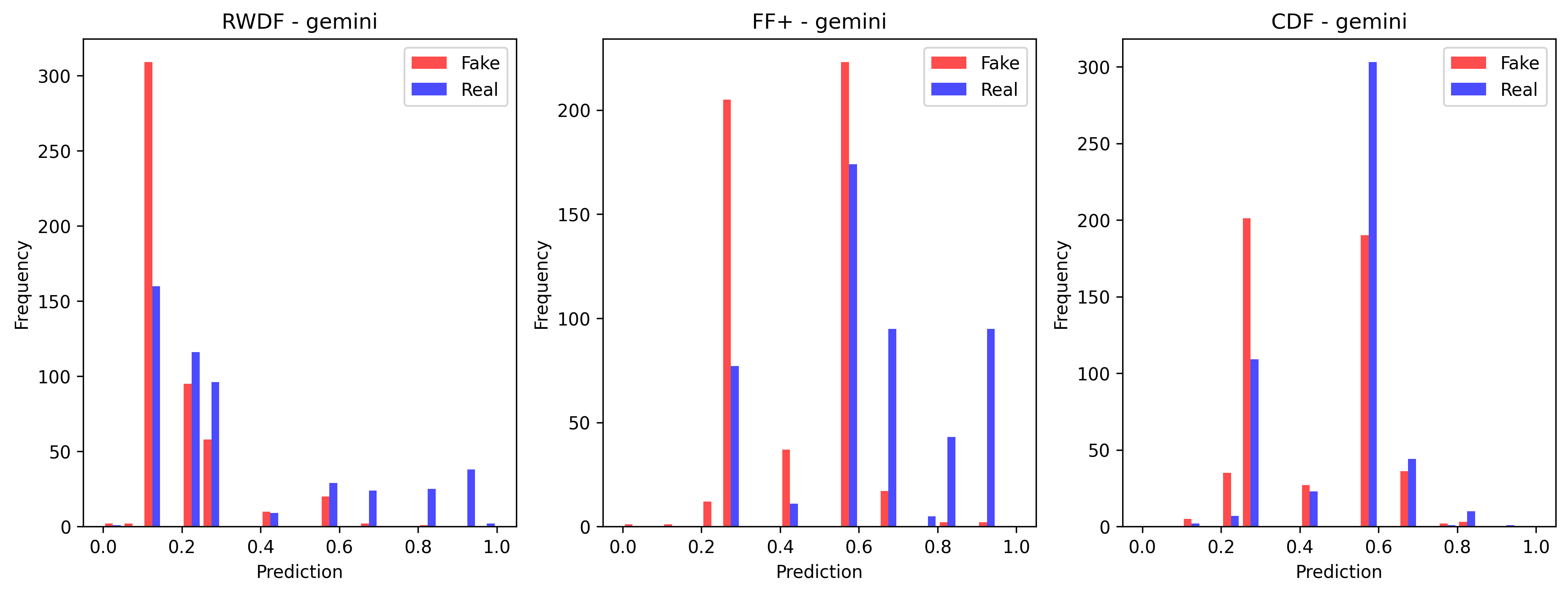}
    \caption{Gemini 2.0 Flash Histogram}
    \label{fig:gemini_flash}
\end{figure*}

\begin{figure*}[h]
    \centering
    \includegraphics[width=\textwidth]{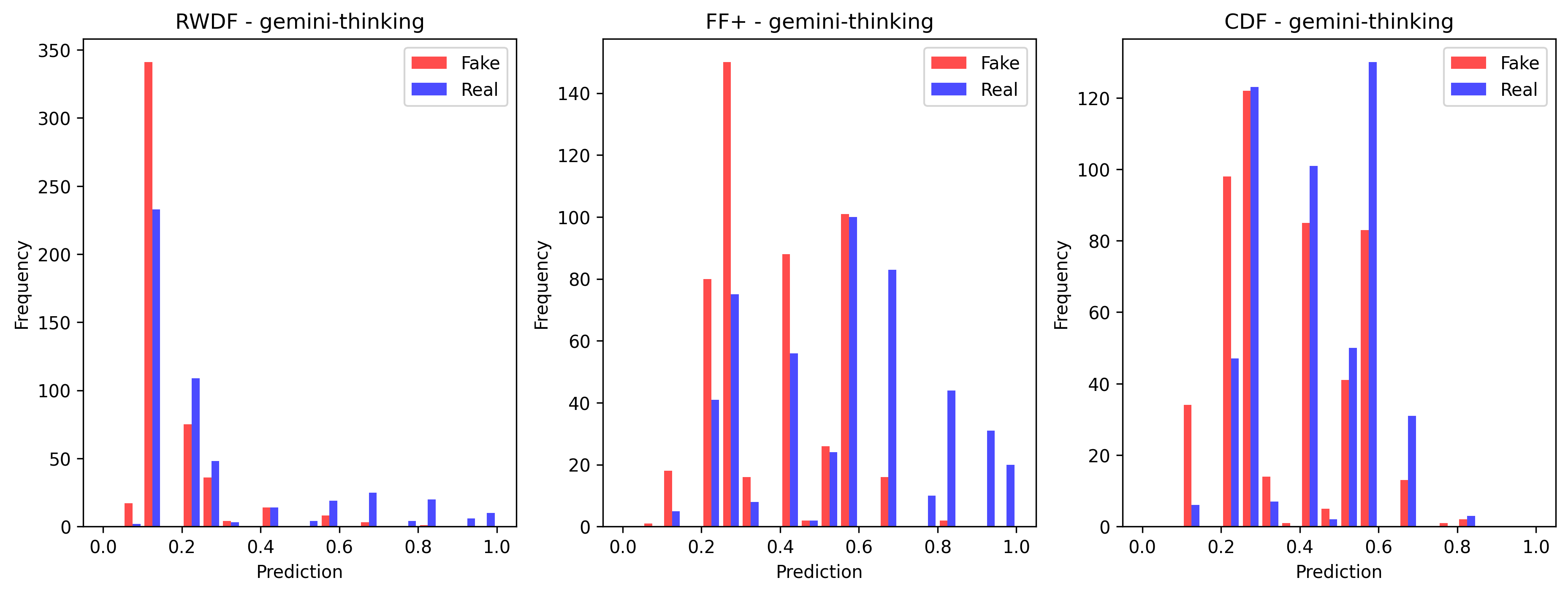}
    \caption{Gemini 2.0 Flash Thinking Histogram}
    \label{fig:gemini_thinking}
\end{figure*}

\begin{figure*}[h]
    \centering
    \includegraphics[width=\textwidth]{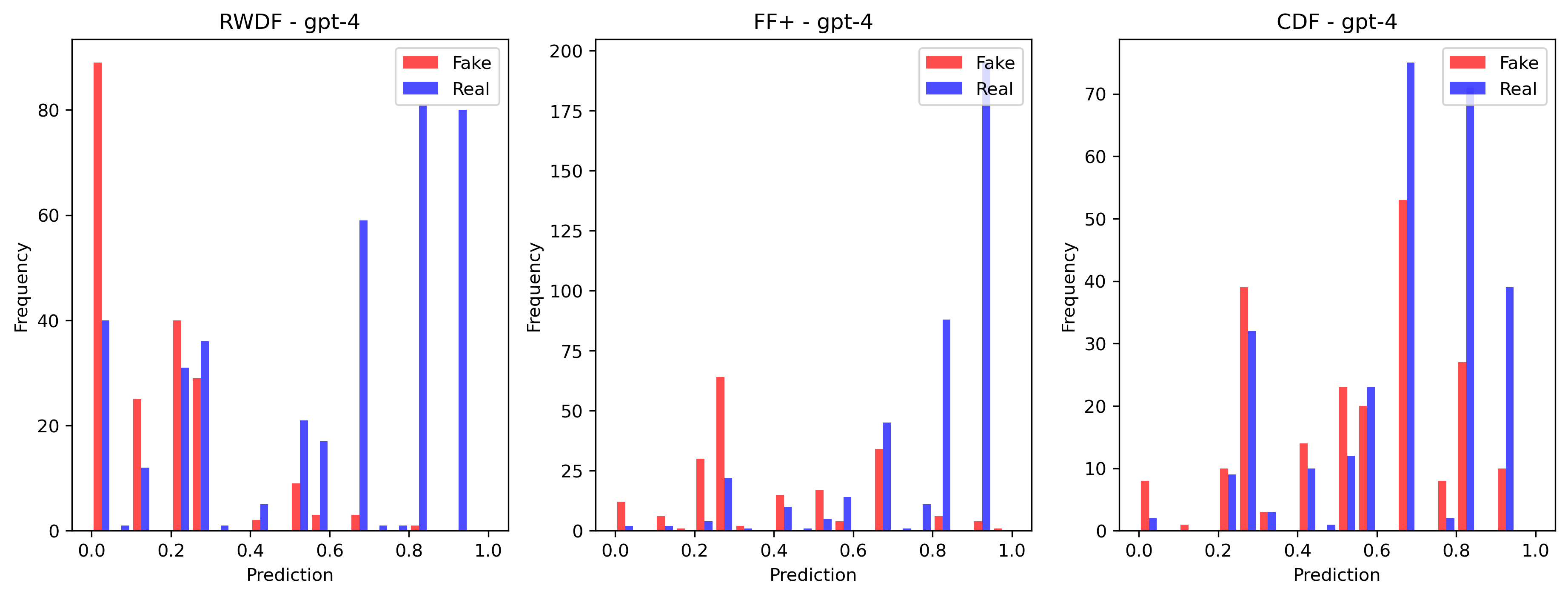}
    \caption{GPT-40 Histogram}
    \label{fig:gpt_40}
\end{figure*}

\begin{figure*}[h]
    \centering
    \includegraphics[width=\textwidth]{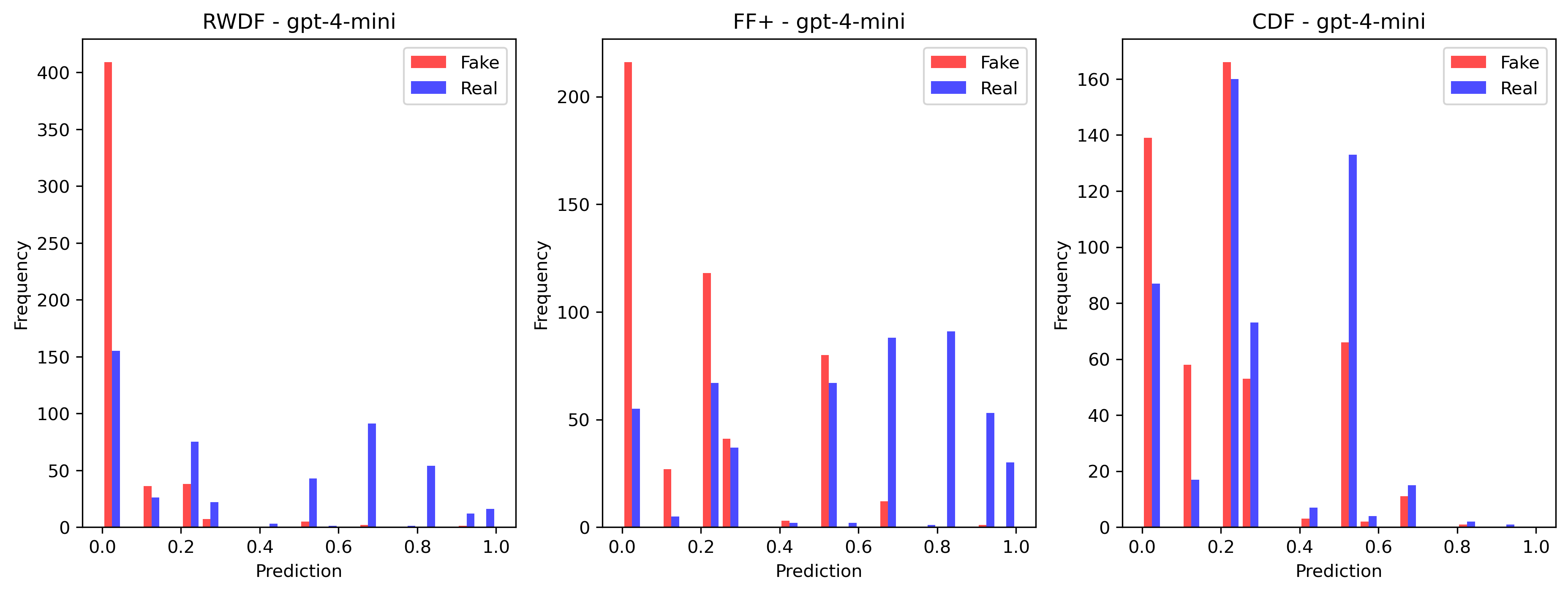}
    \caption{GPT-40 Mini Histogram}
    \label{fig:gpt_40_mini}
\end{figure*}

\begin{figure*}[h]
    \centering
    \includegraphics[width=\textwidth]{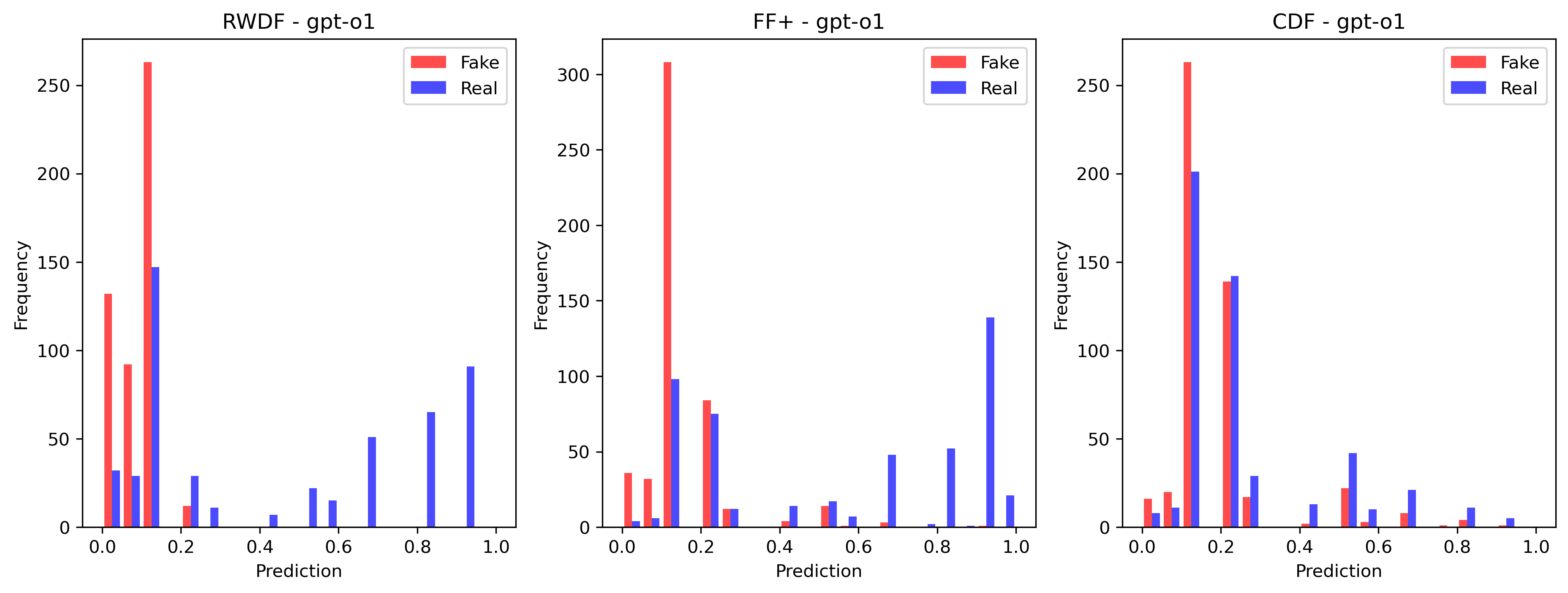}
    \caption{GPT-01 Histogram}
    \label{fig:gpt_01}
\end{figure*}

\begin{figure*}[h]
    \centering
    \includegraphics[width=\textwidth]{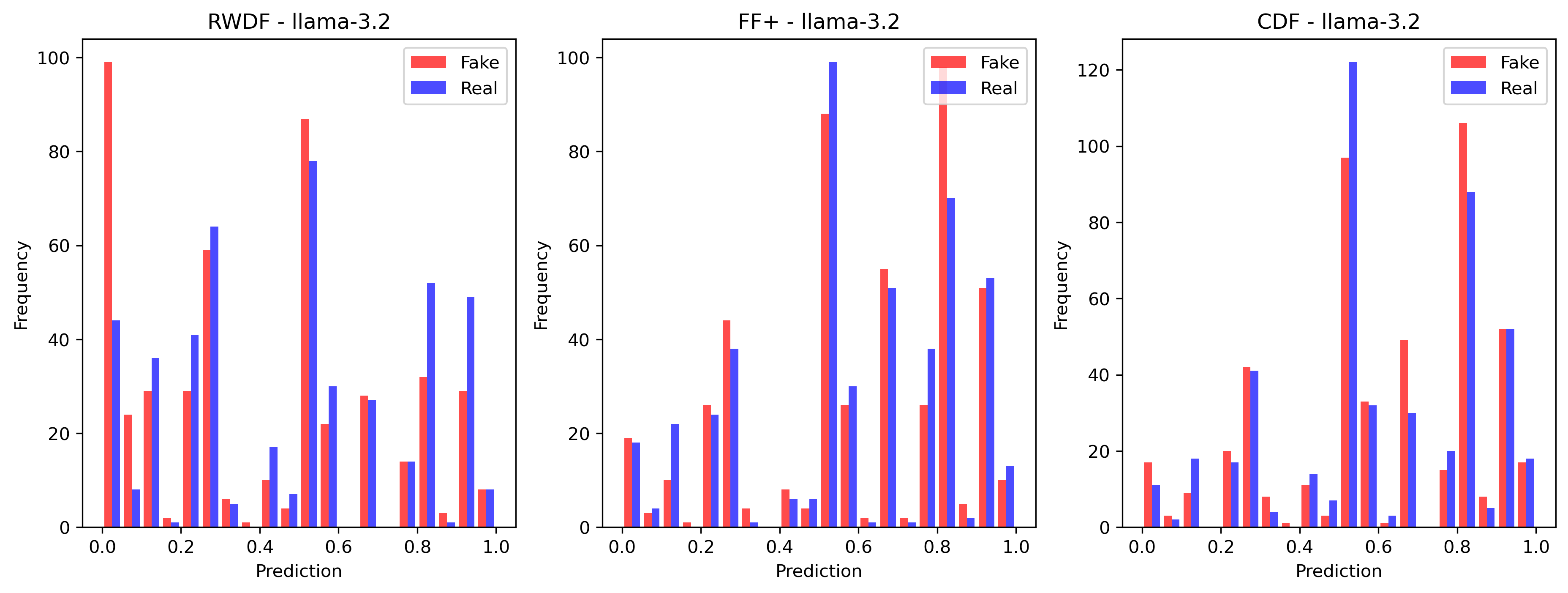}
    \caption{Meta LLaMA 3.2 11B Vision Histogram}
    \label{fig:meta_llama_11b}
\end{figure*}

\begin{figure*}[h]
    \centering
    \includegraphics[width=\textwidth]{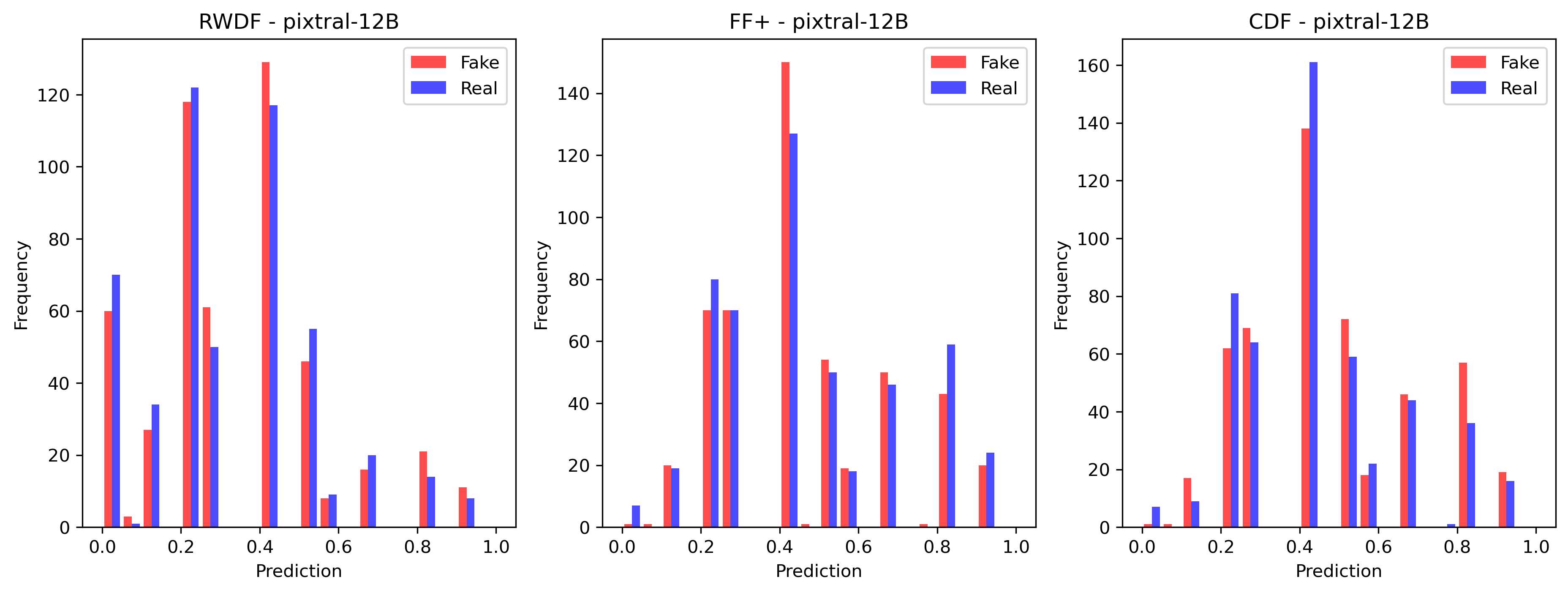}
    \caption{MistralAI Pixtal 12B Histogram}
    \label{fig:mistral_pixtal}
\end{figure*}

\begin{figure*}[h]
    \centering
    \includegraphics[width=\textwidth]{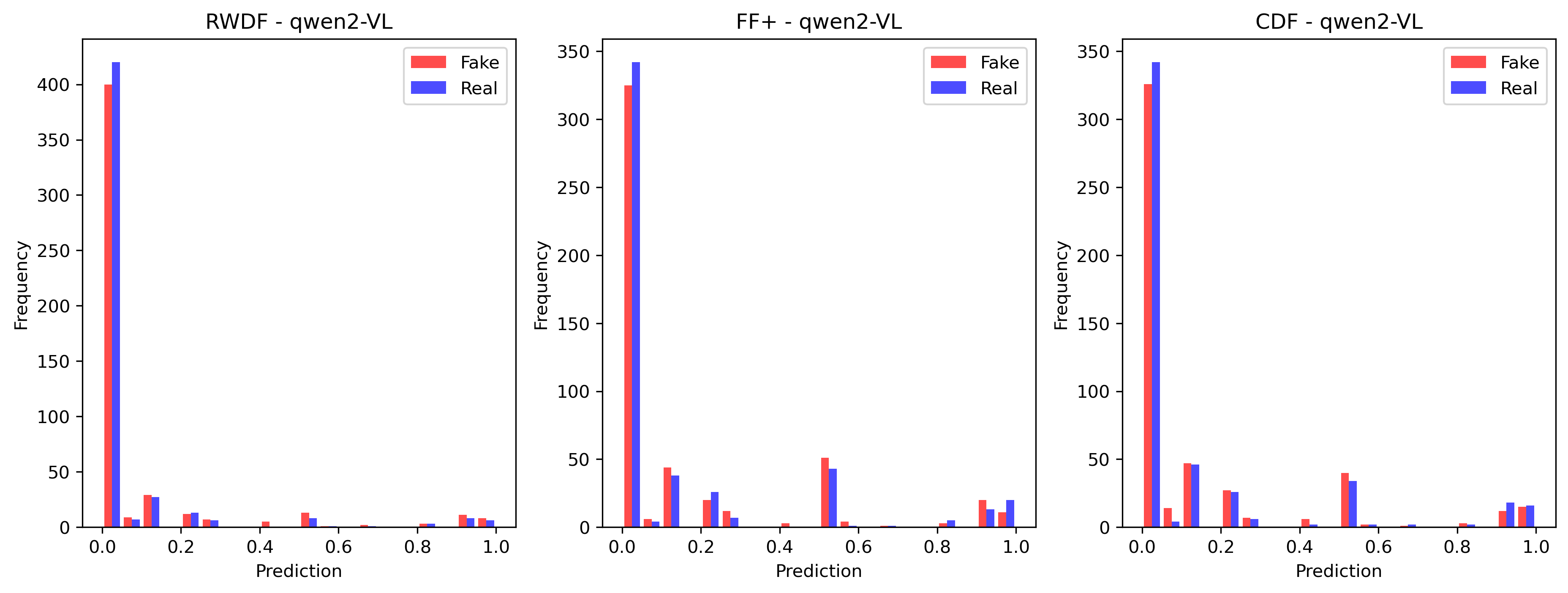}
    \caption{Qwen Qwen2 VL 7B Instruct Histogram}
    \label{fig:qwen_7b}
\end{figure*}

\begin{figure*}[h]
    \centering
    \includegraphics[width=\textwidth]{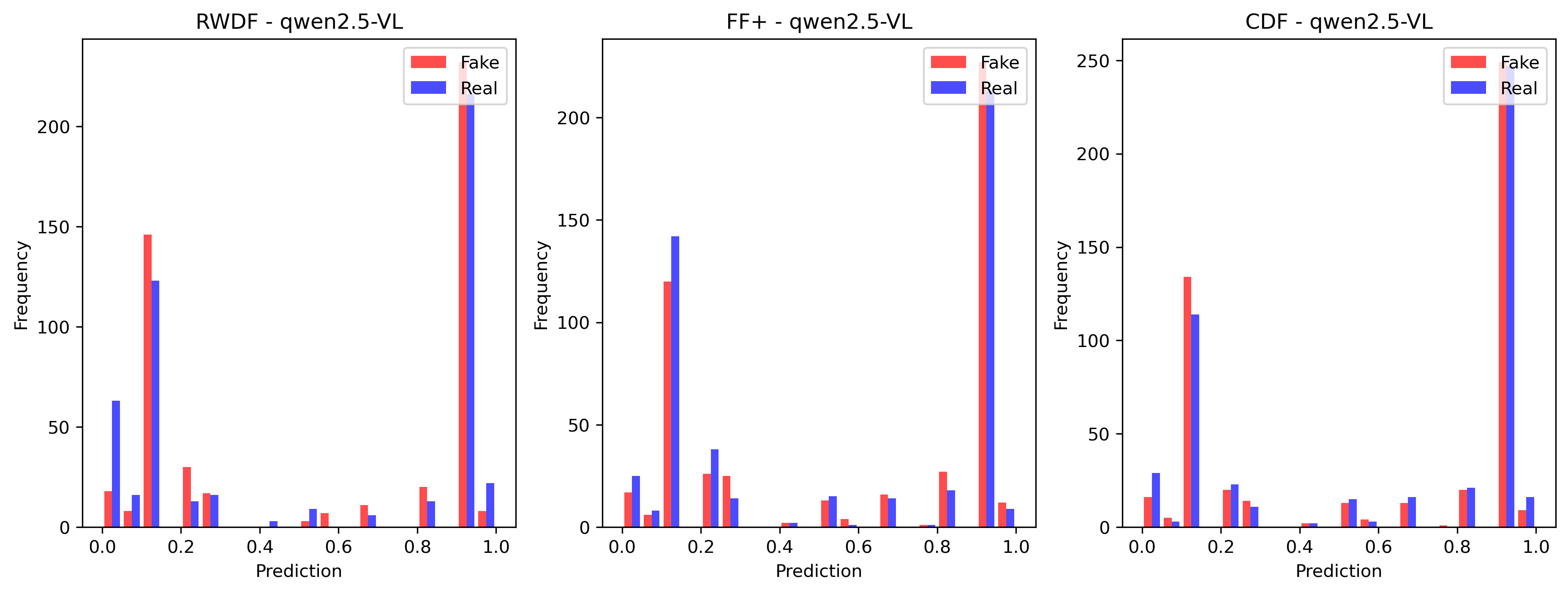}
    \caption{Qwen Qwen2.5 VL 7B Instruct Histogram}
    \label{fig:qwen_7b_25}
\end{figure*}

We plot the histogram of the output for each of our models. 
\begin{itemize}
    \item Histogram for Claud 35 Haiku: See Figure \ref{fig:claud_35_haiku}.
    \item Histogram for Claud 37 Sonnet: See Figure \ref{fig:claud_37_sonnet}.
    \item Histogram for Deepseek AI Janus Pro 1B: See Figure \ref{fig:deepseek_1b}.
    \item Histogram for Deepseek AI Janus Pro 7B: See Figure \ref{fig:deepseek_7b}.
    \item Histogram for Gemini 2.0 Flash: See Figure \ref{fig:gemini_flash}.
    \item Histogram for Gemini 2.0 Flash Thinking: See Figure \ref{fig:gemini_thinking}.
    \item Histogram for GPT-40: See Figure \ref{fig:gpt_40}.
    \item Histogram for GPT-40 Mini: See Figure \ref{fig:gpt_40_mini}.
    \item Histogram for GPT-01: See Figure \ref{fig:gpt_01}.
    \item Histogram for Meta LLaMA 3.2 11B Vision: See Figure \ref{fig:meta_llama_11b}.
    \item Histogram for MistralAI Pixtal 12B: See Figure \ref{fig:mistral_pixtal}.
    \item Histogram for Qwen Qwen2 VL 7B Instruct: See Figure \ref{fig:qwen_7b}.
    \item Histogram for Qwen Qwen2.5 VL 7B Instruct: See Figure \ref{fig:qwen_7b_25}.
\end{itemize}

\section*{Data and Code Availability}
The datasets and code supporting this work are released by Scam.AI and are available from
the Scam.AI Research portal, \url{https://scam.ai/research}~\cite{scamai_research}. Document-
and PDF-forgery resources are additionally maintained at Trustable, \url{https://trustable.ai}~\cite{trustable_ai}.
These are the canonical, maintained locations for the resources associated with this paper and
supersede any prior ad hoc distribution.

\bibliographystyle{plain}
\bibliography{egbib}
\end{document}